\documentclass[10pt, a4paper]{article}
\usepackage{lrec}
\usepackage{graphicx}
\usepackage{tabularx}
\usepackage{soul}

\usepackage{epstopdf}
\usepackage[utf8]{inputenc}

\usepackage{hyperref}
\usepackage{xstring}

\usepackage{color}

\usepackage{amsmath}
\usepackage{url}
\usepackage{xfrac}
\usepackage{xspace}
\usepackage{amssymb}
\usepackage{tabularx}
\usepackage{multirow}
\usepackage{comment}
\usepackage{siunitx}
\usepackage{amsmath}
\usepackage{flushend}
\usepackage{caption}
\usepackage{multicol}
\usepackage{multirow}
\usepackage{algorithm}
\usepackage[noend]{algpseudocode}
\usepackage[colorinlistoftodos]{todonotes}
\usepackage{comment}
\usepackage{enumitem}
\usepackage{breqn}
\usepackage{tabularx}
\usepackage{arydshln}
\usepackage{graphicx}
\usepackage{makecell}
\usepackage{caption}
\usepackage{subcaption}
\usepackage{url}
\usepackage{color,soul}
\usepackage{xcolor}
\usepackage{booktabs}

\newcommand{\citet}{\newcite}
\newcommand{\citep}{\cite}

\def\reals{\mathbb{R}}

\newcommand\blfootnote[1]{%
  \begingroup
  \renewcommand\thefootnote{}\footnote{#1}%
  \addtocounter{footnote}{-1}%
  \endgroup
}

\newcommand{\club}{$^\clubsuit$}
\renewcommand{\diamond}{$^\diamondsuit$}
\newcommand{\spade}{$^\spadesuit$}

\title{Learning Word Ratings for Empathy and Distress \\ from Document-Level User Responses}

\name{João Sedoc*\club,
Sven Buechel*\textsuperscript{\dag}\spade,
Yehonathan Nachmany\diamond,
Anneke Buffone\diamond, and
Lyle Ungar\diamond
}

\address{
\club Johns Hopkins University,
\spade Friedrich-Schiller-Universität Jena,
\diamond University of Pennsylvania\\jsedoc@jhu.edu, sven.buechel@uni-jena.de, yoninachmany@gmail.com, buffonea@sas.upenn.edu, ungar@cis.upenn.edu\\}

\abstract{
Despite the excellent performance of black box approaches to modeling sentiment and emotion, lexica (sets of informative words and associated weights) that characterize different emotions are indispensable to the NLP community because they allow for interpretable and robust predictions. Emotion analysis of text is increasing in popularity in NLP; however, manually creating lexica for psychological constructs such as empathy has proven difficult. This paper automatically creates empathy word ratings from document-level ratings. The underlying problem of learning word ratings from higher-level supervision has to date only been addressed in an {\em ad hoc} fashion and has not used deep learning methods. We systematically compare a number of approaches to learning word ratings from higher-level supervision against a Mixed-Level Feed Forward Network (MLFFN), which we find performs best, and use the MLFFN to create the first-ever empathy lexicon. We then use Signed Spectral Clustering to gain insights into the resulting words. The empathy and distress lexica are publicly available at: \url{http://www.wwbp.org/lexica.html}.\\
\newline  }

\begin{document}

\maketitleabstract

\section{Introduction}
\label{sec:intro}
Deep learning, applied to ever larger datasets, has led to  large improvements in performance in sentiment and emotion analysis. In  light of this development, lexica, lists of words and associated weights for a particular affective variable, which used to be a key component for feature extraction \citep{Mohammad17starsem}, may seem obsolete. However, this is far from the truth.
\blfootnote{* These authors contributed equally to this work. Sven Buechel designed the MLFFN model (Section \ref{sec:methods}). João Sedoc, along with Yoni Nachmany, was responsible for conducting experiments (Section \ref{sec:experiments}) and clustering analysis (Section \ref{sec:ssc}). Experimental design and supervision of the implementation of the algorithms were done jointly by both first authors.}
\blfootnote{\dag Work partially conducted as a visiting researcher at the University of Pennsylvania.}

Lexica can be used as features to improve performance for sentence-level emotion prediction even in advanced neural architectures  \citep{Mohammad17wassa,DeBruyne19arxiv}. Word ratings are also often used to refine pre-trained embedding models for specific tasks \citep{Yu17emnlp,Khosla18coling}.
But much more importantly, word ratings are relatively cheap to acquire and have been found to be robust across domains and even languages, regarding their translational equivalents \citep{Leveau12,Warriner13}. This gives lexica a pivotal role for processing under-resourced languages.
Perhaps most importantly, using lexica allows for interpretable models since the resulting document-level predictions can be easily broken down to the words within it. This gives lexica an important role for building justifiable AI and addressing related ethical challenges~\citep{clos2017towards}.
Interpretability is also crucial for  NLP use in other academic disciplines such as psychology, social science, discourse linguistics, and the digital humanities, where understanding the nature of ``constructs'' (as psychologists call them) such as emotions is far more important than making accurate predictions \citep{schwartz2013toward,eichstaedt2015psychological,pennebaker2011,liu2016analyzing}.

While lexica for many kinds of emotion already exist (see Section~\ref{sec:existinglexica}), there is no such resource for empathy despite its growing popularity in the NLP community \citep{Khanpour17,Buechel18emnlp}.
Hand-curated lexica for empathy are difficult to create in part because there is no clear set of words that can accurately distinguish empathy from self-focused distress. The gold standard for discerning these is an emotion rating scale by \citet{batson1987distress}. This scale is a collection of emotion words (e.g., compassionate, tender, warm) that could serve as a rudimentary lexicon, but it contains many words that are rarely used (e.g., ``perturbed''), and many words that can take on meanings that are far from empathy (e.g, ``warm'', ``tender''). These word-based scales have shown good reliability for self-report of these emotional states, but would make poor guides for a proper lexicon of empathy.

In this paper, we construct the first empathy lexicon. Specifically, we learn ratings for \textit{two kinds of empathy}---empathic concern (feeling for someone) and personal distress (suffering with someone)---for words given existing document-level ratings from the recently published Empathic Reactions dataset \citep{Buechel18emnlp}. We first train a model to predict document-level empathy in a regular supervised set-up and then "invert" the resulting model to derive word ratings. We conclude with an in-depth analysis of the resulting resource.

\section{Related Work}
\label{sec:related}

\subsection{Lexica for Psychological Quantities}
\label{sec:existinglexica}
The notion of describing (part of) a word's meaning, such as the emotion typically associated with it, in terms of numerical ratings has a long tradition in psychology, dating back at least to \citet{osgood_measurement_1957}.
Today, many sets of word ratings exist, covering numerous constructs and languages, particularly relating to sentiment and emotion. Early work in NLP was mostly focused on positive-vs.-negative resources such as SentiWordNet and VADER \citep{baccianella_sentiwordnet_2010,hutto_vader_2014}. In contrast, resources from psychologists tend to focus on \textit{valence} and \textit{arousal} (or other representations of affective states \citep{Ekman92}). In particular, this includes the Affective Norms for English Words (ANEW; \citep{Bradley99anew}) which have been adopted to many languages \citep{Redondo07,Montefinese14}, and their extension by \citet{Warriner13}. 
Such lexica have recently became popular in NLP \citep{Wang16acl,sedoc2017predicting,mohammad_obtaining_2018,buechel-hahn-2018-emotion}. Lexica also exist for many other constructs, including concrete/abstractness, familiarity, imageability and humor \citep{brysbaert_concreteness_2014,yee_valence_2017,engelthaler_humor_2017}. Yet, noticeably, an empathy lexicon is missing.

Psychologists use such lexica either for content analysis, most noticeably using the Linguistic Inquiry and Word Count (LIWC) lexica \citep{tausczik2010psychological}, or as  controlled stimuli for experiments, e.g.,  on language processing  and memory \citep{hofmann_affective_2009,monnier_semantic_2008}. Applications of lexica in NLP have been discussed in Section \ref{sec:intro}.

Whereas most lexica are created manually, there is an extensive body of work on learning such ratings automatically (see \citet{kulkarni2019depth} for a survey). Early work focused on deriving scores through linguistic patterns or statistical association with a small set of seed words \citep{Hatzivassiloglou97,Turney03}. More recent approaches almost always rely on word embeddings \citep{hamilton_inducing_2016,li_inferring_2017,Buechel18naacl}. This line of work is predominantly based on \textit{word-level} supervision. In contrast, we learn word ratings from document-level ratings.

\subsection{Empathy and Distress in Psychology}

Empathic emotions, ``reactions of one individual to the observed experiences of another'' \citep{davis1983}, often in response to their need or suffering, is complex and controversial, with luminary scientists both arguing for the benefits of empathy \citep{dewaal2009} and
``against empathy''~\citep{bloom2014}.
Empathy has been linked to a multitude of positive outcomes, from volunteering~\citep{batson1997}, to charitable giving~\citep{pavey2012}, and even longevity~\citep{poulin2018}, but it can also cause the empathizing person increased stress~\citep{buffone2017} and emotional pain~\citep{chikovani2015}.
In this paper, we build lexica for two distinct types of  {\em state} (momentary) empathy, empathic concern and personal distress, which are based on the subscales of Interpersonal Reactivity Index (IRI) questionnaire~\citep{davis1980}. The scale creator defines these as follows: {\bf Empathic Concern} assesses ``other-oriented'' feelings of sympathy and concern for unfortunate others and {\bf Personal Distress} measures ``self-oriented'' feelings of personal anxiety and unease in tense interpersonal settings. For conciseness, we will use the terms ``empathy'' and ``distress'' to refer to this pair of constructs throughout the paper.

\subsection{Empathy and Distress in AI}

Most previous work in language-centered AI for empathy has been conducted with a focus on speech and especially spoken dialogue. Conversational agents, psychological interventions, and call center applications have been addressed particularly often \citep{Mcquiggan2007,Fung16,Perez-Rosas17,Alam17}.
In contrast, studies addressing empathy in written language are surprisingly rare.
\citet{Abdul2017icwsm}, in contrast, focus on \textit{trait} empathy, a temporally more stable personal attribute. In particular, they studied the detection of ``pathogenic empathy'', marked by self-focused distress, a potentially detrimental form of empathy associated with health risks, in social media language using a wide array of features, including $n$-grams and demographic information.

\citet{Khanpour17} present a corpus of messages from online health communities which has binary empathy annotations on the sentence-level. They report an .78 F-Score using a CNN-LSTM. The corpus, however, is not publicly available.
In contrast, \citet{Buechel18emnlp} recently presented the first publicly available gold standard dataset supported by proper psychological theories. The dataset consists of responses to news articles and scales for empathy and distress between 1 and 7. They collected empathy and distress ratings from the writer of an informal message using a sophisticated annotation methodology  borrowed from psychology.
In this contribution, we build upon their work by using their document-level ratings to predict word labels.

\subsection{Lexicon Learning from Document Labels}

Few studies address learning word ratings based on document-level supervision. However, those studies (described in detail below) focus on their particular application rather than addressing the underlying, abstract learning problem (formalized in Section \ref{sec:methods}). As a result, previously proposed methods have not been quantitatively compared.

In an early study,~\citet{mihalcea2006corpus} computed the \textit{happiness factor} of a word type as the ratio of documents labeled ``happy'' to all blog posts it occurs in. Labels were given by the blog users. The resulting lexicon was used to estimate user happiness over the course of an average 24-hour day as well as a seven-day week.
\citet{Rill12} independently came up with a very similar approach for identifying the evaluative meaning of adjectives and adjective phrases (\textit{absolutely fantastic} vs. \textit{just awful}) based on a corpus of online product reviews. Since the individual reviews come with a one-to-five star rating, the evaluative meaning of an adjective or phrase was computed as the average rating of all reviews it occurs in (Mean Star Rating, see Section \ref{sec:methods}). This approach was later adopted by \citet{Ruppenhofer14} who found that it works quite well for classifying quality and intelligence adjectives into intensity classes (\textit{excellent} vs. \textit{mediocre} and \textit{brilliant} vs. \textit{dim}, respectively).
Another related approach was proposed by \citet{Mohammad12}, who used hashtags in Twitter posts as distant supervision labels of emotion categories, e.g., \textit{\#sadness}. Word ratings were then computed based on pointwise mutual information between word types and emotion labels.

The above methods all derive word labels directly using relatively simple statistical operations. From this group, we selected the Mean Star Rating approach for experimental comparison (Section \ref{sec:experiments}), as it expects numerical document labels, in line with the later employed empathy gold standard (Section \ref{sec:lex}).

Note that these contributions are distinct from pattern-based approaches, e.g., presented by \citet{Hatzivassiloglou97}, who distinguish positive and negative words based on their usage pattern with particular conjunctions: ``A \textit{and} B'' implies that A and B have the same polarity whereas  ``A \textit{but} B'' implies opposing polarity. Such approaches are not considered here because they base lexicon learning on linguistic usage patterns instead of document-level supervision and hence rely on large quantities of raw text.

In another line of work, \citep{Sap14emnlp} address the task of modeling user age and gender in  social media. They showed that by training a linear model with Bag-of-Words (BoW) unigram features, the resulting feature weights can effectively be interpreted as word-level ratings.
In a later study~\citet{Preotiuc16} employed the same method to create a valence and arousal lexicon based on annotated Facebook posts.
This is the second baseline method we used in our evaluation; Technical details are given in Section \ref{sec:methods} (Regression Weights).

In a recent study,~\citet{Wang17} present a three step approach to infer word polarity.
Based on a Twitter corpus with hashtag-derived polarity labels, they (1)  apply the method of~\citet{Mohammad12} to generate a first set of word labels (see above). Those ratings are used (2) to train sentiment-aware word embeddings. The embeddings are then used  (3) as input to a classifier which is trained on a set of seed words to predict the final word ratings. In essence, this is a semi-supervised approach because the last step requires word-level gold data and does not address the problem at hand.

\begin{figure}[t]
    \centering
    \includegraphics[width=.4\textwidth]{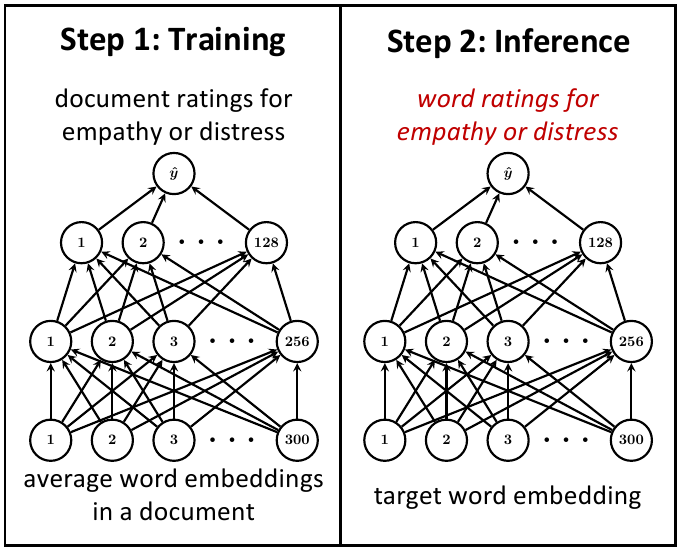}
    \caption{
    Schematic illustration of the MLFFN: {\bf Step 1} Train a model to predict a document  rating from the embedding of the document text; {\bf Step 2} "Invert" the trained model to compute a word rating for each word embedding.
}
    \label{fig:schema}
\end{figure}

\section{Methods}
\label{sec:methods}
This section formalizes the learning problem we address, describes the three baseline methods we compare against and the Mixed-Level Feed Forward Network, and concludes with a brief discussion. Signed Spectral Clustering, which we use for qualitative interpretation of our resulting empathy lexicon, is described in Section \ref{sec:lex}.

\paragraph{Problem Statement.} We address the problem of learning word ratings for an arbitrary lexical characteristic based on gold labels of the same characteristic, but for a higher linguistic level (see  Figure \ref{fig:schema}). For example, how can one learn \textit{word}-level polarity ratings based on \textit{document}-level polarity gold labels?
More formally, let $\mathcal{W}=\{w_1, w_2, ...w_n\}$ denote a set of words with corresponding gold labels $Y^w = (y^w_1, y^w_2, ..., y^w_n)$. Let $\mathcal{D} = \{d_1, d_2, ..., d_m\}$ be a set of higher level linguistic units with corresponding gold labels $Y^d = (y^d_1, y^d_2,...,y^d_m)$. Those linguistic units can be anything from phrases over paragraphs to whole books, yet for conciseness we will refer to those units as \textit{documents}. Our problem is to predict $Y^w$ given $\mathcal{W}$, $\mathcal{D}$, and $Y^d$.
To the best of our knowledge, this is the first contribution ever formulating this as an abstract learning problem (rather than looking at concrete applications in isolation) and studying it in a systematic manner---the baseline methods have so far not been compared against each other. We now proceed by introducing methods for solving this problem.

\paragraph{\textsc{Mean Star Rating}. }
Following \citet{Rill12}, we predict $y^w_i$ by averaging the gold labels of documents $d_j$ in which the word $w_i$ occurs. We denote the set of documents containing $w_i$ as $D(w_i)$. Hence this baseline method can be described as follows:

\begin{equation}
    \hat{y}^w_i = \frac{1}{|D(w_i)|} \sum_{d_j \in D(w_i)} y^d_j.
    \label{eqn:meanstarrating}
\end{equation}

\paragraph{\textsc{Mean Binary Rating.}} As previously mentioned, \citet{mihalcea2006corpus} created lexica for happiness and sadness from binary labels. To apply this method to numerical document labels (as present in the Empathic Reactions dataset; see Section \ref{sec:lex}) we first apply a median split (documents labels below (above) the median are recorded as 0 (1)). Subsequently, we calculate Mean Binary Rating using the same equation as for Mean Star Rating (Equation \ref{eqn:meanstarrating}) thus showing the resemblances between \citet{mihalcea2006corpus} and \citet{Rill12}.

\paragraph{\textsc{Regression weights.}}
Following \citep{Sap14emnlp}, this baseline method learns word ratings by fitting a linear regression model with Bag-of-Words (BoW) features.

First, consider a linear regression model for predicting  document ratings $Y^d$. In general, such a model is given by
 \begin{equation*}
     \hat{y}^d_i = a_0 + \sum_{j \in \text{\it features }} a_j \cdot x_j,
 \end{equation*}
where $a_0$ denotes the intercept and $a_j$ and $x_j$ represent weight and value for feature $j$, respectively.

Using a BoW approach, \textit{relative frequency} of a word in a document is often used as features. Except for the intercepts, the linear model can conversely be interpreted as computing the weighted average of \textit{all weight terms} $a_j$, the relative term frequency then being the weighting factor. With this interpretation in mind, a linear BoW model aligns perfectly with a lexicon-based approach to achieve document-level prediction, with feature weights corresponding to word ratings (see \citet{Sap14emnlp} for a more detailed explanation). Hence the above equation can be rewritten as

 \begin{equation*}
     \hat{y}^d_i = a_0 + \sum_{w_j \in \mathcal{W}} \hat{y}^w_j \cdot \mathrm{rf}(w_j, d_i),
 \end{equation*}
where $\mathrm{rf}(w_j, d_i)$ denotes the relative frequency of word $w_j$ in document $d_i$. Thus, by fitting the model to predict document ratings, we learn word ratings, simultaneously. In practice, ridge regression is used for fitting the model parameters (word ratings) thus introducing $\ell_2$ penalization to avoid overfitting.

\paragraph{\textsc{Mixed-Level Feed Forward Network}.}

We learn a Feed Forward Network (FFN; illustrated in Figure \ref{fig:schema}) on the document-level using a neural BoW approach with an external, pre-trained embedding model. By training the FFN on this task, it implicitly learns to map points of the embedding space to gold labels, which we then exploit for predicting word level ratings.

In general, a Feed Forward Network consists of an input layer $a^{(0)} \in \reals^{\mathrm{dim}}$ followed by multiple hidden layers with activation,
\begin{equation*}
a^{(l+1)} = \sigma (W^{(l+1)}a^{(l)} + b^{(l+1)}  ),
\end{equation*}
where $W^{(l+1)}$, $b^{(l+1)}$ denote weights and biases of layer $l+1$, respectively, and $\sigma$ is a nonlinear function.
Since we predict numerical values (document-level ratings), the activation on the output layer $a^{(out)}$, where $out$ is the number of non-input layers, is given by the affine transformation
\begin{equation*}
   \hat{y}^d =  a^{(out)} = W^{(out)}a^{(out-1)} + b^{(out)}.
\end{equation*}

For fitting the model parameters, consider a pre-trained embedding model such that $\mathrm{vec}(\omega) \in \mathbb{R}^{\mathrm{dim}}$ denotes the vector representation of a word $\omega$. This would be either the learned representation of $\omega$ or a zero vector of length $\mathrm{dim}$ if $\omega$ is not in the embedding model.
We can now train the model to predict the document gold ratings $Y^d$ using a gradient descent-based method. For a document $d_i$, the embedding centroid of tokens present in $d_i$ is used as input $a^{(0)}(d_i)$ . That is,
\begin{equation*}
a^{(0)}(d_i) = \frac{1}{\mathrm{len}(d_i)}\sum_{\omega \in d_i}\mathrm{vec}(\omega),
\end{equation*}
where $\mathrm{len}(d_i)$ is the number of tokens in $d_i$. Embeddings are not updated during training.

Until this point, the described model is quite similar to deep averaging networks (DAN) proposed by \citet{Iyyer15acl} in that it is a Feed Forward Network that predicts document labels from embedding centroid features.
What differs is that we used the model to predict \textit{word} labels, once it is fit to predict the document labels $Y^d$. Conceptually, by fitting the model parameters, the FFN learns to map points of the (pre-trained) embedding space to points in the label space of $Y^d$. But using the same embedding model, we can also represents words $w_i \in \mathcal{W}$, the ones we want to predict labels for, within the same feature space.
Moreover, note that per our problem definition, word and document labels populate the same label space. Hence, we can predict $y_i^w$ by feeding $\mathrm{vec}(w_i)$ into the FFN without any further adjustments. Since the FFN can predict both word \textit{and} document labels, we call this model Mixed-Level Feed Forward Network (MLFFN). \footnote{For the MLFFN, our method of using individual words to derive ratings is mathematically equivalent to SHapley Additive exPlanations (SHAP) \citep{lundberg2017unified}.}

\paragraph{Hyperparameters and Implementation. }
The implementation of Mean Star Rating and Mean Binary Rating is straightforward and requires no further details.
For Regression Weights, we used the same setup as \citet{Sap14emnlp}, as implemented in the Differential Language Analysis Toolkit (DLATK, \citep{Schwartz17emnlp}).

For MLFFN, we built on the implementation\footnote{
Implementation available along with the Empathic Reactions dataset; see Footnote \ref{footnote:empathic_reactions}.} and hyperparameter choices \citet{Buechel18emnlp} used for the Empathic Reactions dataset. Thus, MLFFN has two hidden layers (256 and 128 units, respectively) with ReLU activation. The model was trained using the Adam optimizer \citep{Kingma15} with a learning rate of $10^{-3}$ and a batch size of 32. We trained for a maximum of 200 epochs, and applied early stopping if the performance on the validation set did not improve for 20 consecutive epochs. We applied dropout with probabilities of $.2$ and $.5$ on input and dense layers, respectively. Moreover $\ell^2$ regularization of $.001$ was applied to the weights of dense layers. Keras~\citep{chollet2015keras} was used for implementation.

\paragraph{Discussion of Model Properties.}

Mean Star Rating, Binary Star Rating, and Regressions Weights learn exclusively from the available document-level gold data. In contrast, one of the major advantages of the MLFFN is that it builds on pre-trained word embeddings, thus implicitly leveraging vast amounts of unlabeled text data. For our experiments we use publicly available embeddings which are trained on hundreds of billions of tokens. MLFFN is also more flexible than Regression Weights, since it can learn nonlinear dependencies between relative word frequencies of a document and its gold label.

Another major advantage of the MLFFN model relates to the set of words that gold labels can be predicted for. Whereas Mean Star Rating, Mean Binary Rating, and Regression Weights are conceptually limited to words which occur in the gold data, MLFFN can predict ratings for any word for which embeddings are known. In practice, this implies that with our approach empathy ratings for \textit{millions} of word types can be induced.

\section{Experiments}
\label{sec:experiments}

\begin{table}[t]
    \centering
\begin{tabular}{l|ccc|c}
\toprule
& \multicolumn{3}{c}{\textbf{Intr}.} & \textbf{Extr.} \\
 \textbf{Method} & V & A & D
\\ \midrule
Mean Binary Rating & .31  & .18  & .11  & .07  \\
Regression Weights & .36      & .22   & .13  & .09   \\
Mean Star Rating   & .39   & .22  & .14   & .09\\ MLFNN  & {\bf .64} & {\bf .45} & {\bf .50} & {\bf .18}\\                     \bottomrule           
\end{tabular}
    \caption{Evaluation of lexicon learning methods in Pearson correlation. Left column group: intrinsic evaluation on emotion corpus and lexicon (VAD = Valence-Arousal-Dominance); Right column group: Extrinsic evaluation on Twitter corpus annotated with person-level empathy.}
    \label{tab:eval}
\end{table}

We next conduct a systematic comparison of the above approaches.
The best evaluation strategy would require having both document-level and word-level ratings for empathy. One could then train the models on the former and test the performance in predicting the later, possibly using resampling  to get a distribution of scores. However, this option is not available, since the difficulty of acquiring empathy word ratings is exactly the point of this paper.\footnote{
    Another seemingly obvious evaluation strategy would be to  predict document-level ratings from derived word-level lexica using the empathic reactions dataset in a cross-validation setup. However, we found that this approach has two major drawbacks. First, rather then validating the resulting word ratings directly, this strategy constitutes a down-stream (extrinsic) evaluation, similar to what we propose in Section \ref{sec:experiments.extrinsic}. Second, we found empirically that this approach lacks statistical power to distinguish between methods due to an insufficient number of examples.
    }
    
We adopt two alternative approaches: First, in place of empathy, we rely on other affective variables, namely, valence, arousal, and dominance (VAD), for which both document and word ratings are available. The assumption here is that performance results for VAD are transferable to empathy. Second, we use the Empathic Reactions\footnote{\label{footnote:empathic_reactions}\url{https://github.com/wwbp/empathic_reactions}} dataset to create one lexicon for each method under consideration. We then use it to predict \textit{trait}-level empathy ratings, thus testing the generalizability of the resulting lexica to other domains as well as from state to trait empathy (see Section \ref{sec:related}).

\subsection{Intrinsic Evaluation with  Emotion Data}

We use the following gold standards for evaluation: Document-level supervision is provided by EmoBank~\citep{buechel2017emobank}\footnote{\url{https://github.com/JULIELab/EmoBank}}, a large-scale corpus manually annotated with emotion according to the psychological Valence-Arousal-Dominance scheme. EmoBank contains ten thousand sentences with multiple genres and has annotations from both writer and reader emotion.
Word-level supervision to test against comes from the well-known affective norms  (psychological valence, arousal, and dominance) dataset collected by \citet{Warriner13} containing 13,915 English word types.

We fit all four models on EmoBank and evaluate against the word ratings by \citet{Warriner13} using 10-fold cross-validation.
For word embeddings we used off-the-shelf Fasttext subword embeddings~\citep{mikolov2018advances}.\footnote{ \url{https://dl.fbaipublicfiles.com/fasttext/vectors-english/crawl-300d-2M-subword.zip}.} The embeddings are trained with subword information on Common Crawl (600B tokens).
Performance will be measured in terms of Pearson correlation between predicted and gold labels. As shown in Table \ref{tab:eval}, the MLFFN by far outperforms Mean Binary Rating, Mean Star Rating and Regression Weights, the latter three being roughly equal. This is most probably, because the MLFFN builds on top of a pre-trained embedding model, thus leverage vast amounts of unlabeled data in addition to the document-level supervision.

\subsection{Extrinsic Evaluation}
\label{sec:experiments.extrinsic}

To evaluate the lexica created using each of the underlying methods, we applied them to measure \textit{trait}-level empathy dataset. We validate our empathy lexica by showing that they predict personal-level empathy traits on another dataset which collected trait-level empathy questionnaires and  users' Facebook posts. \citet{Abdul2017icwsm} used this dataset to predict person trait-based pathogenic empathy. Here, instead we aggregate the empathy survey results of Facebook users recruited via Qualtrics. We filtered users to include those who had posted at least five times in the last 30 days and had at least 100 lifetime posts.  The survey included an integrated app grabbing participants’ Facebook posts. In total there are 2,405 users with  1,835,884 Facebook posts after filtering non-English posts (see \citet{Abdul2017icwsm} for further dataset details).

The lexica were employed in a very simple fashion: For each user, we computed the weighted average of the empathy scores of the the words they used across all tweets. Relative frequency was used as weighting feature.
As shown in Table \ref{tab:eval}, the performance is generally much poorer, indicating the change of domain, the more difficult task of inferring \textit{trait}-level empathy from state-level ratings, and the overall reduced performance of purely lexicon-based approaches.\footnote{
  While 0.18 may seem low, our results are similar to those from  \citet{Abdul2017icwsm} who use a regression model with LDA topics trained on Facebook posts.}
Nevertheless, our findings are consistent with intrinsic results. The MLFFN widely outperforms the other approaches (having about twice as high performance figures), the latter being roughly similar in performance.

\section{The Empathy Dictionary}
\label{sec:lex}

\begin{table}[t]
    \centering
    \small
\begin{tabular}{llrr}
\toprule
              &             &  \textbf{Empathy} &  \textbf{Distress} \\
\midrule
\textbf{High Empathy} & lukemia &     6.90 &      5.09 \\
              & lakota &     6.70 &      4.88 \\
              & healing &     6.60 &      4.90 \\
              \midrule
\textbf{Low Empathy} & joke &     1.10 &      1.52 \\
              & worrying &     1.10 &      3.93 \\
              & wacky &     1.10 &      1.49 \\
              \midrule
\textbf{High Distress} & inhumane &     4.07 &      6.55 \\
              & dehumanizes &     5.46 &      6.40 \\
              & mistreating &     4.85 &      6.31 \\
              \midrule
\textbf{Low Distress} & somehwere &     1.82 &      1.05 \\
              & dunno &     1.31 &      1.05 \\
              & guessing &     1.38 &      1.06 \\
\bottomrule
\end{tabular}
\caption{Illustrative examples from final lexicon: highest/lowest ranking words for empathy and distress.}
    \label{tab:empathydistresswords}
\end{table}

\begin{figure}[t!]
    \centering
    \includegraphics[width=.42\textwidth]{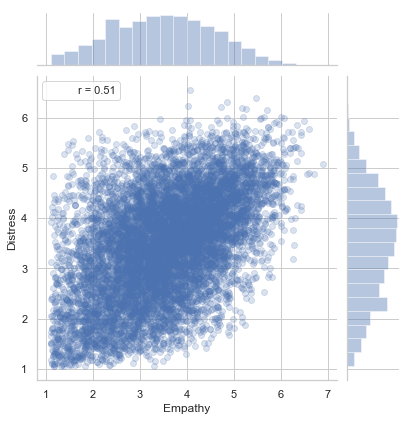}
    \caption{Label distribution in our empathy lexicon.}
    \label{fig:dist}
\end{figure}

\begin{table*}[tb!]
    \centering
    \small
    \begin{tabular}{ll}
\toprule
\textbf{High Empathy} &  $\bullet$ grieve, grieving, loss, prayers, grief, heartbroken, losses, deppression, condolences, widowed \\
 & $\bullet$  wounds, wounded, scars, heal, blisters, trauma, wound, heals, bleeding, fasciitis \\
&  $\bullet$ duckworth, salama, mansour, santiago, gilbert, fernandez, braves, vaughn, colonialism, crowe \\
 &  $\bullet$ minneapolis, neighborhoods, detroit, chicago, charlotte, cincinnati, brisbane, angeles, atlanta, drescher \\
\midrule
\textbf{Low Empathy}&    $\bullet$ fool, clueless, dumbass, idiotic, lazy, stupidity, morons, idiot, idiots, dumb \\
 &  $\bullet$ bother, slightest, anything, else, nobody, anybody, any, nothing, anyone, himself \\
 & $\bullet$  loser, bs, moron, dingus, maniac, buffoon, ffs, loon, crap, psycho \\
 & $\bullet$  wacky, bizarre, odd, creepy, weird, unnerving, masochistic, freaks, unusual, strange \\
\midrule
 \textbf{High Distress}& $\bullet$  homicide, killings, murdered, massacre, murdering, homicides, genocide, murder, murderers, killed \\
 & $\bullet$  brutalized, assaulted, raped, bullied, tormented, harassed, detained, molested, reprimanded, beaten \\
&  $\bullet$  horrific, witnessed, retched, wretched, atrocious, awful, horrid, foul, shoddy, unpleasant \\
 & $\bullet$  horrifying, terrifying, harrowing, overdoses, suicides, deaths, suicide, gruesome, devastating, tragedy \\
\midrule
\textbf{Low Distress} & $\bullet$  dunno, guessing, guess, gues, probably, assuming, maybe, clue, bet, assume \\
 &  $\bullet$ wont, knowlegde, alot, doesnt, isnt, wasnt, ahve, dont, didnt, exempt \\
& $\bullet$  sort, lot, bunch, sorts, type, whatever, plenty, depending, types, range \\
& $\bullet$  intact, stays, rememeber, keeping, keeps, always, kept, vague, rember, stay \\
 \bottomrule
     \end{tabular}
    \caption{Signed Spectral Clustering results for qualitative analysis.}
    \label{tab:empathyclusters}
\end{table*}

The final empathy dictionary consists of the predictions of the MLFFN from the last experiment, which we then adjusted using log min-max rescaling
into the interval $[1, 7]$ for consistency with \citet{Buechel18emnlp}. We restricted ourselves to words which appear in the Empathic Reaction dataset and did not make use of the ability of the MLFFN to predict ratings for all word of the embedding model (left for future work). This was done to ensure interpretability of word ratings relative to their usage in the corpus (achieved via clustering analysis in Section \ref{sec:ssc}).

\subsection{Dataset Description}
Our final lexicon consists of 9,356 word types (lower-cased, non-lemmatized, including named entities and spelling errors) each with associated empathy and distress ratings. For illustration Table~\ref{tab:empathydistresswords}, lists the highest and lowest ranking words for each construct (empathy and distress).
High-empathy words contain many named entities that experience or cause suffering  making a reader feel empathic (e.g., \textit{lukemia} or \textit{lakota}). This is likely because the Empathic Reactions corpus used news stories to evoke empathy in subjects who then referred to those named entities for expressing their feeling  (Section \ref{sec:ssc} provides an estimate of the total number of named entities in the lexicon). Low-empathy words, on the other hand, are often ones used for ridiculing, hence expressing a lack of empathy (\textit{joke}, \textit{wacky}). High-Distress words contain predominantly adjectives, nouns, and participles which can be used to characterize abusive behaviour (\textit{inhumane}, \textit{mistreating}) thus causing personal distress in readers when taking the perspective of the affected entity. Interestingly, low-distress words do not seem to display any  clear pattern, making us suspect that personal distress should be addressed in terms of a unipolar rather than a bipolar scale.

Uni- and bi-variate distribution of empathy and distress scores is displayed in Figure \ref{fig:dist}. As can be seen, both sets of labels are fairly close to a normal distribution.
Empathy and distress word-level ratings display only a moderate Pearson correlation of $r \approx .51$ which confirms that both are distinct constructs as already indicated by the qualitative analysis above. It is also highly consistent with Empathic Reactions where \citet{Buechel18emnlp} found $r\approx .45$ between both sets of \textit{document-level} ratings.

\subsection{Clustering Analysis}
\label{sec:ssc}

To assess the face validity of the lexica, we partition the lexica into groups (clusters) of words that are semantically similar and simultaneously have similar ratings.  Straightforward clustering does not take the ratings into account, and is less interpretable.
We use the Signed Spectral Clustering (SSC) algorithm to cluster words that are similar semantically and in their ratings~\citep{sedoc2017predicting}. Weighted edges are added between words such that words of similar empathy have positive connections and those of differing empathy are negative (see \citet{sedoc2017semantic} for precise mathematical formulation).
SSC minimizes the cumulative edge weights cut within clusters versus between clusters, while simultaneously minimizing the negative edge weights within the clusters, thus pulling words of similar empathy or distress into the same clusters and pushing those that differ away.  We follow the method used by \citet{sedoc2017predicting}. In terms of linguistic analysis, the resulting clusters can help us describe the \textit{language of empathy} by providing us with semantic groups of words which are high or low on either of the two empathy scales, i.e., allowing us to answer questions such as ``what kind of words do people use when they feel empathic?''

As seen in Table~\ref{tab:empathyclusters}, the clusters of words for high and low empathy and for high distress are strikingly well illustrated.
There are many clusters of topics around situations where people feel empathy. Furthermore, there are lists of different negative emotions. The lists that are all places and people names are less useful obviously for psychological analysis. However, these lists are places where bad things happen, and people to whom bad things happen, which is useful for predictive models. Usable lexica must be interpretable, SSC allows us not only to give words and ratings, but also, groups of high magnitudes. These allow domain experts then to analyze and possibly modify the lexica.

\subsection{Named Entities}
Motivated by the observation that groups of named entities (NEs) play a (perhaps surprisingly) prominent role in our lexica, we used the clusters to derive an estimate of the number of NE entries in the resulting dataset.
The authors manually labeled clusters which belong to either of the following classes of NEs: person (names), organization (including geopolitical entity), date or time, number (including units of measurement), and punctuation. If a cluster predominantly contains entries of one NE type, then all words within this cluster were counted as belonging to this category. The categorization was done independently for the empathy and distress lexica. The clusters tend to be consistent regarding NE types, as illustrated in Table \ref{tab:empathyclusters}. However, both false positives (FP; non-NE entries in NE clusters) as well as false negatives (FN;  NE entries in non-NE clusters) do occur. This leads to slightly different estimates of the number of NEs in the empathy and distress lexicon, respectively.\footnote{Although both dictionaries comprise the same list of word types, the clusters are split differently leading to deviating FP and FN rates.}

The approximate NE counts are presented in Table \ref{tab:named_entity_counts}. The estimates are largely consistent between the empathy and distress lexica, with only ``Organization'' displaying more pronounced differences. In total, the results indicate the ratio of NE entries in our lexica is roughly 6\%.

\begin{table}[h]
    \centering
    \begin{tabular}{l|rr}
    \toprule
        \textbf{Named Entity Class} & \textbf{Empathy} &\textbf{ Distress } \\
    \midrule
        Person & 136 & 140\\
        Organization & 238  & 165\\
        Date & 52 & 54 \\
        Number & 128 & 117\\
        Punctuation & 48 & 30\\
    \midrule
        \textbf{Sum} &602& 506\\
    \bottomrule
    \end{tabular}
    \caption{\label{tab:named_entity_counts}
    Cluster based estimate of the number of lexicon entries belonging to various classes of named entities.
    }
\end{table}

\section{Conclusion}
This contribution reported on the creation of the first-ever lexica for empathy and distress using a newly proposed model: The Mixed-Level Feed-Forward Network (MLFFN) successfully learns word ratings from document-level ratings by backing out word ratings from a trained neural net, performing substantially better than methods that others have used for lexicon creation. Signed Spectral Clustering was applied to the resulting lexical scores to gain insights into the language of empathy. We look forward to further validating the lexica by using them in predictive models and psychological experiments, and to exploring the extent to which using the SHAP (SHapley Additive exPlanations) \citep{lundberg2017unified} calculations of feature importance for CNNs, RNNs, or Transformers improve lexicon quality over the simple neural nets which we used.

\section{Acknowledgements}
We would like to thank Daphne Ippolito,  Reno Kriz, and the anonymous reviewers for their helpful feedback. This work was partially supported by João Sedoc's Microsoft Dissertation Grant. Sven Buechel was partially funded by the German Federal Ministry for Economic Affairs and Energy (funding line "Big Data in der makroökonomischen Analyse"; Fachlos 2; GZ 23305/003\#002). He would also like to thank his doctoral advisor Udo Hahn, JULIE Lab, for supporting his research visit at the University of Pennsylvania.
This research is supported in part by the Office of the Director of National Intelligence (ODNI), Intelligence Advanced Research Projects Activity (IARPA), via the BETTER Program contract \#2019-19051600005. The views and conclusions contained herein are those of the authors and should not be interpreted as necessarily representing the official policies, either expressed or implied, of ODNI, IARPA, or the U.S. Government. The U.S. Government is authorized to reproduce and distribute reprints for governmental purposes notwithstanding any copyright annotation therein.

\section{Bibliographical References}\label{reference}

\bibliographystyle{lrec}
\bibliography{litfinal}

\end{document}